\documentclass[preprint,12pt]{elsarticle}

\usepackage{lineno}
\usepackage{graphicx}
\usepackage{mathtext}
\usepackage{amssymb,amsmath,latexsym,amsfonts,array,epsfig}
\usepackage{mathrsfs}

\newtheorem{definition}{Definition}[section]
\newtheorem{Th}{Theorem}[section]

\journal{}

\begin{document}

\begin{frontmatter}



\title{Lattice Generalizations of the Concept of Fuzzy Numbers and Zadeh's Extension Principle}


\author{Dmitry Maximov}

\address{Trapeznikov Institute of Control Science Russian Academy of Sciences,
65 Profsoyuznaya str, Moscow;
jhanjaa@ipu.ru; dmmax@inbox.ru}

\begin{abstract}
The concept of a fuzzy number is generalized to the case of a finite carrier set of partially ordered elements, more precisely, a lattice, when a membership function also takes values in a partially ordered set (a lattice). Zadeh's extension principle for determining the degree of membership of a function of fuzzy numbers is corrected for this generalization. An analogue of the concept of mean value is also suggested. The use of partially ordered values in cognitive maps with comparison of expert assessments is considered.
\end{abstract}




\end{frontmatter}


\section{Introduction}
\label{Int}
In the theory of fuzzy sets, carrier sets always have some metric, and membership functions take values in the interval $[0, 1]$. Therefore, we can always compare the set elements by there membership degree. However, we cannot always compare the elements of a lattice, since the lattice is only a partially ordered set (i.e., a poset) in general. Such lattices are considered in the paper as carrier sets and membership scales instead of number sets. The lattices are of interest in the theory of fuzzy sets, since they can serve as scales for expressing expert opinions in more flexible way than usual linear scales \cite{maximov_neuro}, \cite{maximov_22}.

Nevertheless, the concepts of generalization of fuzzy numbers and operations with them, as well as the concept of mean value, are not elaborated yet for lattice-valued variables. We need these concepts if we want to take into account \emph{multiple} experts assessments with some degrees of confidence (membership). We cannot simply use the fuzzy concepts due to the only partial ordering of our values and absence of a division operation in finite case.

Thus, we introduce a lattice generalization of the concept of fuzzy numbers in Sec. \ref{sec1}. In Sec. \ref{sec2}, we also introduce a correction of Zadeh's extension principle of the membership degree for a function of fuzzy numbers in the case of the generalization to lattice values of variables. We consider possible generalizations of the concept of a mean value in Sec. \ref{sec3}.

Cognitive maps with weights and concepts which takes values in a lattice have been considered in \cite{maximov_22}. However, a unique expert assessment was used there for each weight. In this article, in Sec. \ref{sec4}, we demonstrate using multiple expert assessments for each weight in a cognitive map example with degrees of confidence for weights' and concepts' values. Also, we compare the results with various ways of aggregating estimates as an analogues of the mean value concept. We give the necessary definitions used in the text in Appendices, and we conclude the paper in Sec. \ref{con}.

\section{Lattice Generalization of Fuzzy Numbers}
\label{sec1}
We have used general finite lattices in \cite{maximov_neuro}, \cite{maximov_22} as scales for expressing expert assessments instead of the linguistic variable scale of numerical interval subintervals. In such a scale, an element of the lattice corresponds to an expert opinion estimation. Typically, each expert gives an estimate from a very limited set of lattice generators, and the remaining elements of the lattice are obtained as expert estimates as a result of statistical processing. However, if we have several assessments of one expert (or several statistical sets of expert estimates), we get at least one set of lattice elements.  In turn, elements of the set can have some membership degrees.

In the case of fuzzy numbers, the carrier set is a numerical interval and we can always say that there are some set elements with a full membership degree, and some belong to the interval only partially in accordance with the metric. However, it makes no sense to evaluate the degree of membership in a finite carrier set of partially ordered elements without a metric.

\begin{figure}
  \includegraphics[scale=1]{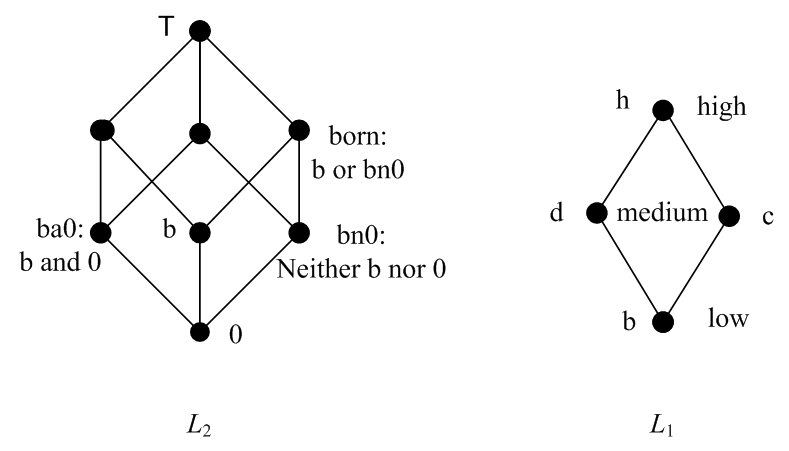}
\caption{Examples of lattices}
\label{fig1}       
\end{figure}

Indeed, let us consider lattices in Fig. \ref{fig1}.
We have no a criterion like a metric in the set $\{d,c\}$ of $L_{1}$ elements or $\{ba0,b,bn0\}$ of $L_{2}$ ones to decide what the element has a higher membership degree. All these elements are incomparable. Thus, it seems reasonable to use a degree of expert confidence instead of the degree of membership.

The difference of these two concepts is that an expert may not be completely sure in any of his assessments. Thus, the confidence function does not have to reach the maximum value on the carrier set unlike the fuzzy numbers. Then, the confidence degrees of partially-ordered assessments may also be partially-ordered --- they are also expert opinions. In this case, the convexity condition of the confidence  function (also, unlike the fuzzy number definition) makes no sense due to only partial ordering and the absence of a metric.

Thus, we define multi-valued sets as a generalization of the fuzzy number concept in the following general way.
\begin{definition}\label{def1}
We refer to a set of pairs
\begin{equation}
A = \{(x,\mu_{A}(x)); x\in L\},
\end{equation}
where
\begin{equation}
\mu_{A}: L \rightarrow M
\end{equation}
is a confidence function of $A$ elements with values in the lattice $M$, as a multi-valued set $A$ in a lattice $L$.
\end{definition}

\section{Lattice Generalization of Zadeh's Extension Principle}
\label{sec2}
Let us remember the extension principle of Zadeh.
\begin{definition}\cite{rut}\label{def2}
Let $X$ be a Cartesian product of crisp sets $X_{1}\times X_{2}\times ... \times X_{n}$. If there is some crisp mapping
\begin{equation}
f: X_{1}\times X_{2}\times ... \times X_{n}\rightarrow Y,
\end{equation}
and also some crisp sets $A_{1}\subseteq X_{1}, A_{2}\subseteq X_{2}, ... A_{n}\subseteq X_{n}$, then the extension principle says that the fuzzy set $B$ formed by the mapping $f$ has the form
\begin{equation}
B = f(A_{1}, ..., A_{n}) = \{(y,\mu_{B}(y)) | y = f(x_{1}, ..., x_{n}), (x_{1}, ..., x_{n})\in X\},
\end{equation}
while
\begin{equation}
\begin{matrix}
\mu_{B}(y) & = & \left\{ \begin{matrix}
\underset{(x_{1}, ..., x_{n})\in f^{-1}(y)}{sup} min\{\mu_{A_{1}}(x_{1}), ..., \mu_{A_{n}}(x_{n})\}, & \mbox{if} \; f^{-1}(y)\neq \emptyset, \\
0, & \mbox{if} \; f^{-1}(y) = \emptyset.
\end{matrix}\right.
\end{matrix}
\end{equation}
\end{definition}
In the lattice-valued case, the max--min operations are replaced by join--meet correspondingly, and this definition becomes not suitable due to some points:
\begin{itemize}
\item In a lattice, in the definition $\mu(y) =\vee(\mu(a)\wedge \mu(b)\wedge ...)$, meets can lead to zero even when all $\mu(a) \neq 0$. And, in general, the more such $\mu(a)$, the more likely the resulting confidence degree is zero: $\mu(y) = 0$.
\item Vice versa, the lattices $L$ and $M$ in Definition \ref{def1} may have an additional structure of a semi-ring or a ring. Then, the addition operation in these structures may be nilpotent. In this case, the resulting confidence degree of the bottom element is maximum: $\mu(0) =\underset{\underset{a\in L}{a+a=0;}}{\vee}(\mu(a)\wedge \mu(a)) = \underset{a\in L}{\vee}(\mu(a)) = \top$, i.e., the lowest estimation is the most confident.
\item Finally, an implication $c = a\rightarrow b$ may be defined in lattices, which is not reduced to arithmetic operations, in contrast to the fuzzy case. If $M = L$ and $\mu(a) = a$, and if there is a separate assessment $\mu(c)$ for the value $c$ of the implication, then $\mu(c) \neq \vee(\mu(a)\wedge \mu(b))$. The situation is the same with all functions in this case.
\end{itemize}
Hence, we propose to use the following formula in the lattice-valued case, which gives the correct result in the case of $M = L$ and $\mu(a) = a$:
\begin{equation}\label{eq6}
\begin{matrix}
\mu_{B}(y) & = & \left\{ \begin{matrix} \underset{x\in f^{-1}(y)}{\vee}(f(\mu_{A}(x))), & \mbox{if} \; f^{-1}(y)\neq \emptyset, \\
0, & \mbox{if} \; f^{-1}(y) = \emptyset.
\end{matrix}\right.
\end{matrix}
\end{equation}
The evident consequence of the formula is that the confidence degree of a function of assessments may be greater than the degree of each of them. Indeed, our confidence in the set of two evaluations is greater than in either of them, if we are not completely sure of either.

\section{Lattice Generalization of Mean Value}
\label{sec3}
A concept of a mean value is needed if we want aggregate different expert assessments in some way. In particular, we may need a mean of lattice-valued analogues of fuzzy numbers. However, we have no a division operation in rings or semi-rings that can be associated with a finite lattice of expert estimates. Thus, there is no a mean concept in this case.

The essence of the mean concept is establishing some common characteristic to a row of values, which can characterize each of them and the whole row at the same time. Thus, we can use a concept of a similarity measure as an analogue of the mean value. There are a lot of such measures between objects that have internal structures \cite{leon}. However, there are only the meet and join operations in the partially-ordered sets without an internal structure, whereas, a lattice of finite sets is our main example in applications.  In this case, we can approach such a concept from the following consideration.

The distance from a point to a set is expressed by the formula $\rho(x,A) = \underset{a\in A}{inf}(\rho(x,a))$ when there is a metric in the universe. We can introduce a measure of difference $d(A,B)$ of two sets $A$ and $B$ of lattice elements following this formula, when the lattice elements are also arbitrary finite sets, and there is no metric:
\begin{equation}\label{eqd}
d(A,B) = \{\underset{a\in A}{\wedge}(b\setminus a)|\forall b\in B\}\cup \{\underset{b\in B}{\wedge}(a\setminus b)|\forall a\in A\}.
\end{equation}
Here, the operation $\setminus$ denotes set difference, and $a$ and $b$ are lattice elements which are also sets of generators.

If we take some type of negation of the measure of difference, we get the definition of a similarity measure as the average of two sets of expert assessments. We can define two types of the mean value: a pessimistic and an optimistic aggregation\footnote{The author is grateful to Dr. O. P. Kuznetsov for the idea of using pessimism and optimism in aggregation}.
\begin{definition}\label{def3}
We refer to the following union of two sets
\begin{equation}
S(A,B) = \{\underset{a\in A}{\vee}(b\wedge a)|\forall b\in B\}\cup \{\underset{b\in B}{\vee}(a\wedge b)|\forall a\in A\},
\end{equation}
as the pessimistic average of two sets $A$ and $B$ of expert assessments.
\end{definition}
\begin{definition}\label{def4}
We refer to the following union of two sets
\begin{equation}
Sopt(A,B) = \{\underset{a\in A}{\wedge}(b\vee a)|\forall b\in B\}\cup \{\underset{b\in B}{\wedge}(a\vee b)|\forall a\in A\},
\end{equation}
as the optimistic average of two sets $A$ and $B$ of expert assessments.
\end{definition}

\begin{figure}
  \includegraphics[scale=0.75]{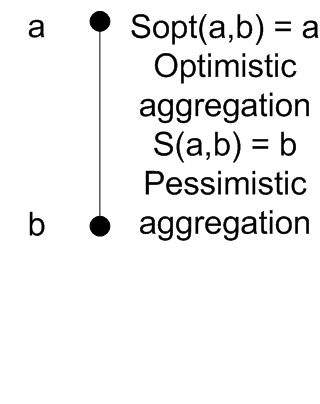}
\includegraphics[scale=1]{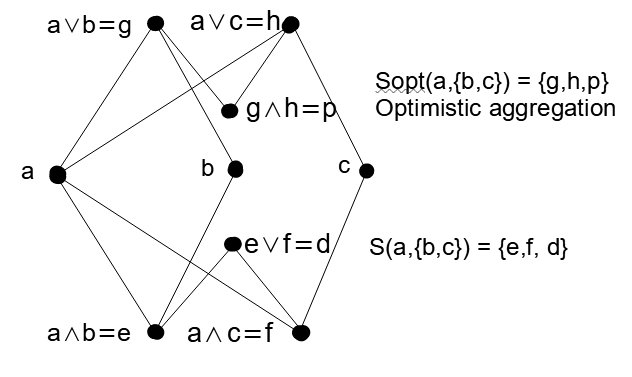}
\includegraphics[scale=1]{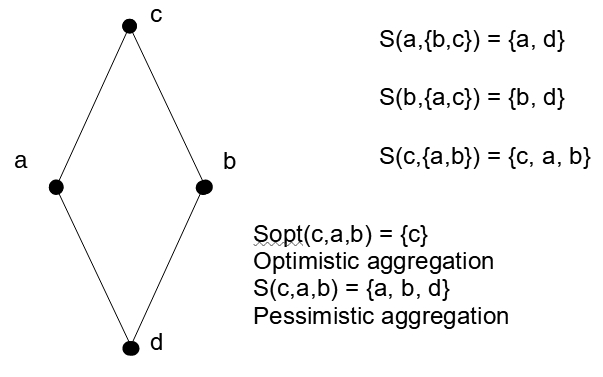}
\includegraphics[scale=0.75]{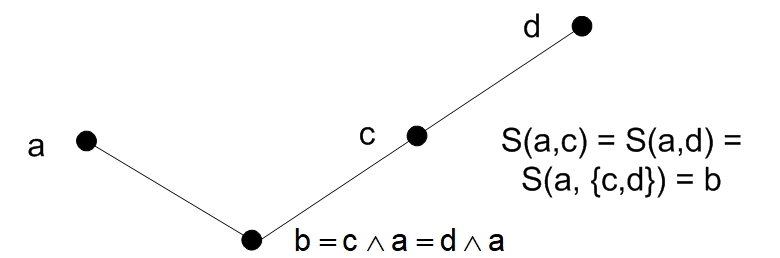}
\caption{Similarity measure examples}
\label{fig2}       
\end{figure}

We get a union of such pairwise aggregations when there are more than two sets in the  arguments of functions $S$ or $Sopt$. One can easy see that the mean value of two sets is also a set with the number of elements from one to the total number of elements in both sets.
Here are some examples to illustrate these concepts in Fig. \ref{fig2}.

The last example in Fig. \ref{fig2} refers to the confidence degree of the mean and shows that different aggregations may have the same confidence value. Hence, the following formula for the confidence degree of the average value is valid ((\ref{eq6}) is used):
\begin{equation}
\begin{matrix}
\mu(S(a^{*})) & = & \left\{ \begin{matrix} \underset{\begin{smallmatrix}a_{i}\in S^{-1}(a^{*}),i=1...n:\\S(a_{1}...a_{n})=a^{*}\end{smallmatrix}}{\vee}S(\mu(a_{1})...\mu(a_{n})), & \mbox{if} \; S^{-1}(a^{*})\neq \emptyset, \\
0, & \mbox{if} \; S^{-1}(a^{*}) = \emptyset.
\end{matrix}\right.
\end{matrix}.
\end{equation}
The formula is suitable for both pessimistic and optimistic aggregation.

\section{Using Partially Ordered Plural Expert Assessments in Cognitive Maps}
\label{sec4}
Here we consider an illustrative model example of the Multi-Valued Cognitive Map (MVCM) from \cite{maximov_22} but with multiple expert assessments, which was originally considered in \cite{groump} in the framework of a Fuzzy Cognitive Map model (FCM). Thus, we consider the example of a Hybrid Energy System combining wind and photovoltaic subsystems; its cognitive map model is depicted in Fig. \ref{fig1}.
\begin{figure}
  \includegraphics[scale=0.8]{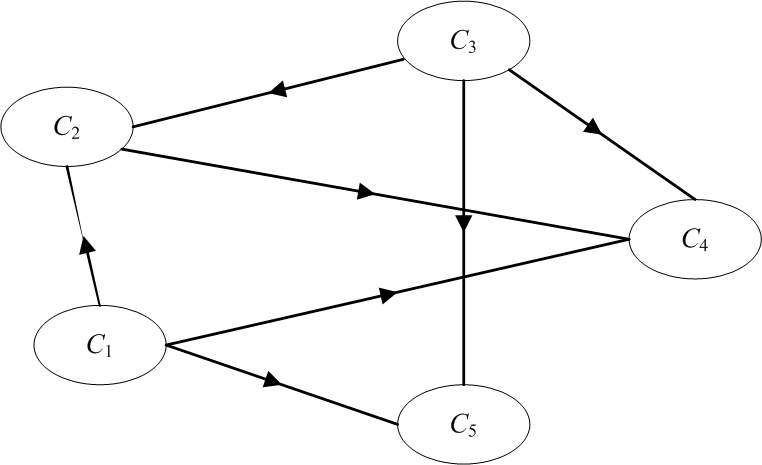}
\caption{The cognitive map of a Hybrid Energy System}
\label{fig3}       
\end{figure}

The model includes the following five concepts:
\begin{itemize}
\item $C_{1}$: sun insolation;
\item $C_{2}$: environment temperature;
\item $C_{3}$: wind;
\item $C_{4}$: PV-subsystem;
\item $C_{4}$: Wind-Turbine-subsystem.
\end{itemize}
In this model there are two energy source decision concepts (outputs), i.e., the two energy sources are considered:
the $C_{4}$: PV-subsystem and the $C_{5}$: Wind-Turbine-subsystem. Concepts $C_{1}$--$C_{3}$ of nature factors influence the subsystems and determine how each energy source will function in this model. The concept's initial values can be obtained from experts' assessments of measurements, which take values in a lattice of linguistic variables. The experts' assessments of concept influence take values also in this lattice. Detailed information for hybrid renewable energy
systems is given in \cite{gould}, \cite{damg}.

The equation that calculates the values of concepts of FCM's with $n$ nodes, can be written in its general form as \cite{glykas}:
\begin{equation}\label{eq11}
A_{i}^{k} = f(d_{1}\sum_{j=1,j\neq i}^{n}w_{ji}A_{j}^{k-1} + d_{ii}A_{i}^{k-1}).
\end{equation}
Here $A_{i}^{k}$ is the value of the concept $C_{i}$ at discrete time $k$, and $d_{ii}$ is a value of self-feedback to node $i$. All values belong to the interval $[0, 1]$, and the function $f$ normalizes its argument up to this interval. Existence and uniqueness of solutions of (\ref{eq11}) in FCM's are proved in \cite{kottas} for some such trimming functions $f$'s. In most applications, $d_{1}$ and $d_{ii}$ are set equal to 1 \cite{groump}.

We use a MVCM model instead of the FCM one, and there is the following general equation for MVCM's in \cite{maximov_22}:
\begin{equation}\label{eq12}
A_{i}^{k} =c_{i}^{k-1}\cdot f_{i}^{k-1}\cdot\bigvee_{j=1}^{n}w_{ji}\cdot A_{j}^{k-1}
\end{equation}
where all weights and concepts take values in a residuated \emph{atomic} lattice $L$, and we use the monoid multiplication of the lattice and  the lattice join, instead of numerical multiplication and sum correspondingly.

The following convergence theorem is proved in \cite{maximov_22}:
\begin{Th}\label{th}
Multi-Valued Cognitive Maps determined by equation (\ref{eq12}) where concepts and weights take values in a finite atomic residuated and integrally-closed lattice $L$ (hence, $L$ is integral), converge under a suitable choice of $c_{i}^{k}$'s and $f_{i}^{k}$'s.
\end{Th}
The following formulas for $c_{i}^{k}$'s and $f_{i}^{k}$'s should be used in order to satisfy the theorem:
\begin{equation}\label{eq13}
f_{i}^{k} = [A_{i}^{k} \vee A_{i}^{k-1}]\leftarrow (\bigvee_{j=1}^{n}w_{ji}\cdot A_{j}^{k})
\end{equation}
\begin{equation}
r_{i}^{k}\leftarrow f_{i}^{k}\leqslant c_{i}^{k} \leqslant 1,
\end{equation}
where
\begin{equation}\label{eq15}
r_{i}^{k} = [A_{i}^{k}\wedge A_{i}^{k-1}]\leftarrow \bigvee_{i=1}^{n}w_{ji}\cdot A_{j}^{k},
\end{equation}
and we can take $c_{i}^{k} = 1$ until the MVCM loops.

In this paper, we take the operation $\wedge$ (meet) as the multiplication in Heyting algebras, i.e., in Brouwer lattices (\ref{a2}) (as in the example in \cite{maximov_22}). Hence, the residuals turn into the lattice implication (\ref{imp}).

However, the lattice of expert assessments may have an additional structure of a ring, and we can use the operations from the structure as the sum and the multiplication in (\ref{eq12})--(\ref{eq15}) instead of the join and the monoid multiplication. The proof of Theorem \ref{th} does not depend on such a replacement.
Here we additionally consider only the sum from the ring of subsets of a finite set (the multiplication in the ring is the lattice meet). In this case the sum is the operation ``excluded or'' or `symmetric difference'' $ a + b = (a\vee b)\setminus (a\wedge b)$ on the lattice $L$, where the lattice elements $a$ and $b$ are considered as sets of generators. Let us note that the sum is nilpotent, i.e., $ a + a = 0$.

So finally  we use the following formulas in our example of MVCM:
\begin{equation}\label{eq16}
C_{i}^{k} =c_{i}^{k-1}\wedge f_{i}^{k-1}\wedge\bigvee_{j=1}^{n}w_{ji}\wedge A_{j}^{k-1}
\end{equation}
or
\begin{equation}\label{eq17}
C_{i}^{k} =c_{i}^{k-1}\wedge f_{i}^{k-1}\wedge\sum_{j=1}^{n}w_{ji}\wedge A_{j}^{k-1}
\end{equation}
for calculating concept values; and
\begin{gather}\label{eq18}
f_{i}^{k} = (\bigvee_{j=1}^{n}w_{ji}\wedge A_{j}^{k})\Rightarrow [A_{i}^{k} \vee A_{i}^{k-1}]
\end{gather}
or
\begin{gather}\label{eq19}
f_{i}^{k} = (\sum_{j=1}^{n}w_{ji}\wedge A_{j}^{k})\Rightarrow [A_{i}^{k} \vee A_{i}^{k-1}];
\end{gather}
and
\begin{gather}\label{eq20}
r_{i}^{k} = (\bigvee_{i=1}^{n}w_{ji}\wedge A_{j}^{k})\Rightarrow [A_{i}^{k}\wedge A_{i}^{k-1}]
\end{gather}
or
\begin{gather}\label{eq21}
r_{i}^{k} = (\sum_{i=1}^{n}w_{ji}\wedge A_{j}^{k})\Rightarrow [A_{i}^{k}\wedge A_{i}^{k-1}]
\end{gather}
for calculating coefficients.

\begin{figure}
  \includegraphics[scale=0.8]{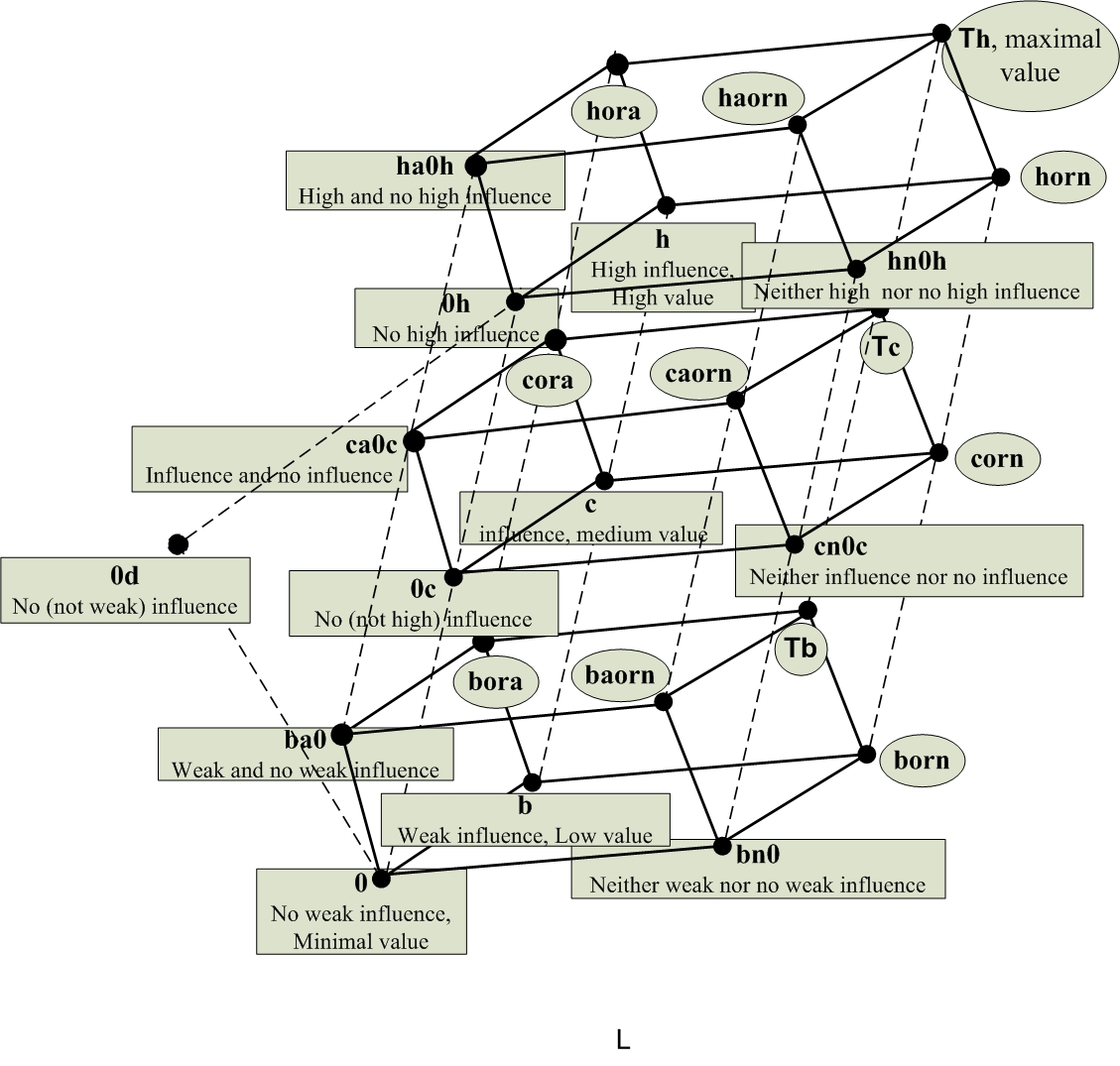}
  \includegraphics[scale=0.5]{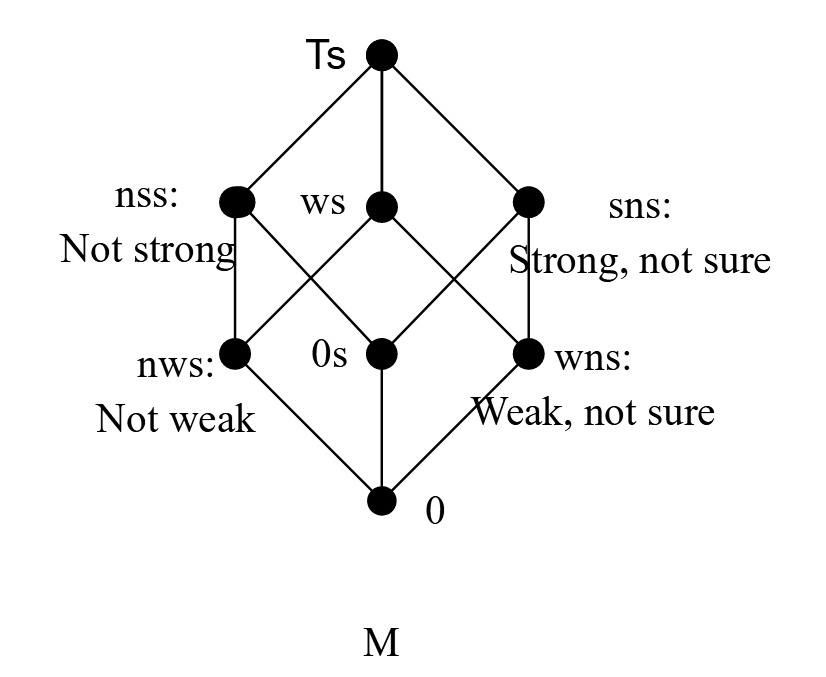}
\caption{The lattice scale of weights' and concepts' assessments --- L and the lattice scale of their confidence estimates --- M}
\label{fig4}       
\end{figure}

We use the following bi-lattice $L$ as the scale of variables \cite{maximov_22}, where $L = L_{1}\times L_{2}$ (lattices $L_{1}$ and $L_{2}$ are depicted in Fig. \ref{fig1}), and the lattice $M$ as the scale of confidence Fig. \ref{fig4}.
We consider the bi-lattice $L$ as the lattice where the unique partial order is generated by atoms $ba0, b, bn0, 0c$, and $0d$. We use two linearly-ordered branches in the lattice $L_{1}$ in order to  regularly obtain the distributive and atomic lattice $L$. Hence, we may use the meet $\wedge$ as the multiplication. The nodes $c$ and $d$ in the $L_{1}$ Fig. \ref{fig1}, corresponding to the medium value, may be interpreted as ``not high'' and ``not weak'' uncertainty degrees (this difference is not reflected in the interpretations in Fig. \ref{fig4}).

A variant of the lattice $L_{2}$ was proposed in \cite{Maximov_Ax} as the interpretation of N. A. Vasil'ev's logic ideas. Vasil'ev has suggested three types of statement: positive, negative and indifferent, instead of only positive and negative. He considered also intermediate types as a hesitation between these main ones. Similarly, we consider here three main uncertainty degrees: $ba0, b, bn0$ (Fig. \ref{fig3}) --- some assessment ``$b$'', the estimation ``$bn0$: Neither $b$ nor 0'', and the estimate ``$ba0$: $b$ and 0''. Similarly, we interpret the generators of sublattices at the levels $c - d$ and $h$ Fig. \ref{fig4}. The assessment, e.g., ``$born$: $b$ or $bn0$'' is the join of $b$ and $bn0$ and can be considered as the hesitation between $b$ and ``Neither $b$ nor 0''. Similarly, ``$bora$: $b$ or $ba0$'' is the join of $b$ and $ba0$ and can be interpreted as the hesitation between $b$ and ``$b$ and 0'' and so on. Then, such hesitations refer to sets of expert assessments, which are ambiguous and contain different subsets of estimates. Thus, we obtain many different variants of uncertainty degrees in assessments, usually obtained statistically when experts use the generators as estimates.

The semantic of the lattice elements is not unique: e.g., the review \cite{pena_sos} contains another semantic of $L_{2}$ elements that is used in \emph{causal} cognitive maps. There, the lattice of the type $L_{2}$ contains eight elements: $+, 0, -$ are generators (they correspond to our $ba0$, $b$ and $bn0$) and $\oplus, \ominus, \pm, a, ?$ are obtained from them  by the following connectives: $\oplus = 0\vee +; \ominus = 0\vee -; \pm = \oplus \vee \ominus; ? = +\vee -\vee 0; a = +\wedge -\wedge 0$ that correspond to $bora$, $born$, etc (our $0$ is $a$ in this notation). Such assessments valuate connections between concepts and, e.g., $A \rightarrow +B$ means ``$A$ excites $B$'', $A \rightarrow -B$ means ``$A$ hurts $B$'', $A \rightarrow 0B$ means ``$A$ is neutral to $B$''; the meaning of connective $\vee$ as in this paper, and  $\wedge$ means ``ambivalent''. Then, the lattice $L_{1}$ distributes values of $L_{2}$ by degrees of influence in the same way as in this article. Perhaps such an intuition, like Vasiliev's, could be more appropriate for the reader, when $+$ means ``positive'', $-$ means ``negative'', $0$ means ``neutral'', and $a$ means ``ambivalent''.

\begin{figure}
  \includegraphics[scale=0.55]{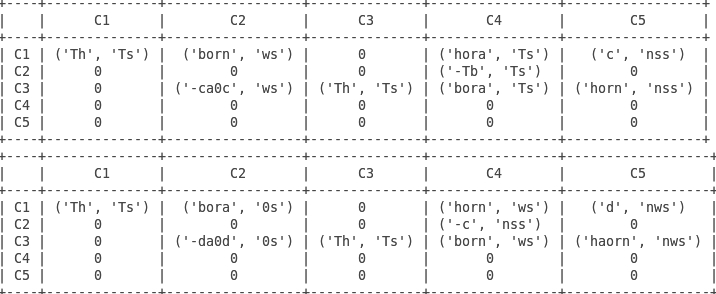}
\caption{Weight matrices of expert opinions with confidence degrees}
\label{fig5}       
\end{figure}

In general, all variables in formulas (\ref{eq16})--(\ref{eq21}) may be multi-valued sets, i.e., may be sets with confidence degrees of their elements. However,  for simplicity, we will use only a set of two assessments and only for weights. Thus, we use two following weight matrices Fig. \ref{fig5}, where the first element in parentheses is the weight, and the second is the degree of confidence of this weight element.

\begin{figure}
  \includegraphics[scale=1.6]{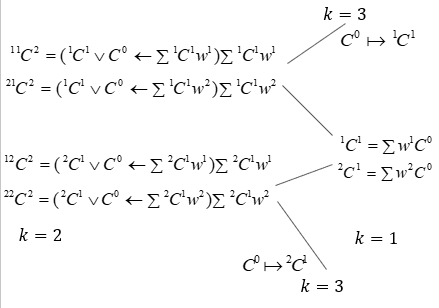}
\caption{The map iterations; at the third step, $C^{0}$ is changed to the corresponding $C^{1}$, and so on}
\label{fig6}       
\end{figure}

During the process of the map iteration, the weights from both these matrices may arise in formulas (\ref{eq16}) and (\ref{eq17}) Fig. \ref{fig6}.
The same process is going for the confidence degrees of the weights and concepts due to (\ref{eq6}). Here, e.g., the variable ${}^{11}C^2$ means the second iteration ($k = 2$) where $w$ is taken from the first matrix, in which the first iteration ${}^{1}C^1$ is also with $w$ from the first matrix.

\begin{figure}
  \includegraphics[scale=0.9]{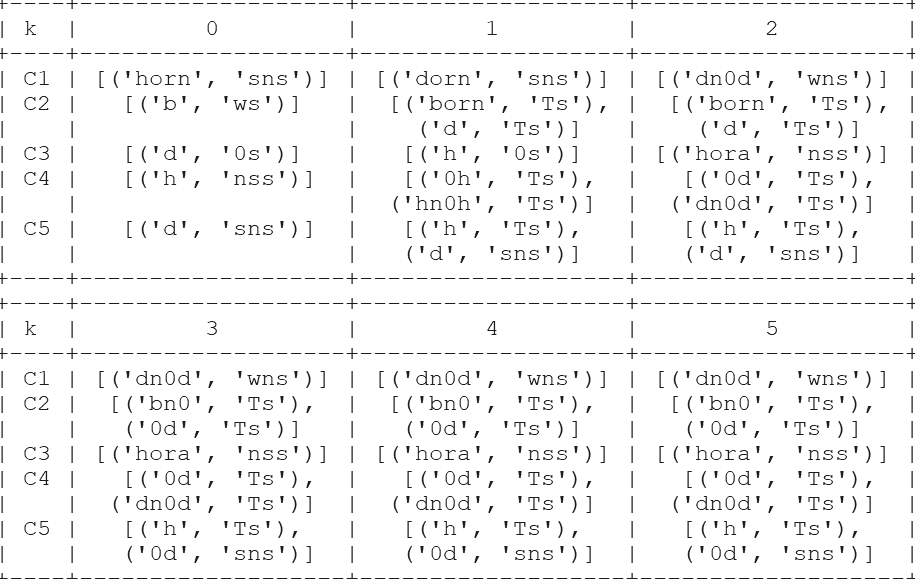}
\caption{The map iterations of separate calculations; insolation decreases, wind increases}
\label{fig7}       
\end{figure}

At the beginning, we present the results of calculations for two matrices separately, i.e., when all variables are lattice elements, not sets Fig. \ref{fig7}\footnote{Here and below, the confidence degrees are calculated according to (\ref{eq6}), where $M \neq L$, and initial concept confidences are determined a priori}.
The first line in each cell corresponds to the opinion of the first expert and behaves just as well as in \cite{maximov_22} --- the wind turbine output rises while the photovoltaic system output falls. However, the second row, corresponding to the second weight matrix, shows the non-physical end result $C_{5}$ --- the output of the wind turbine decreases as the wind increases. Thus, the second expert is not qualified enough.


\subsection{Results with join in formulas}
\begin{figure}
  \includegraphics[scale=0.9]{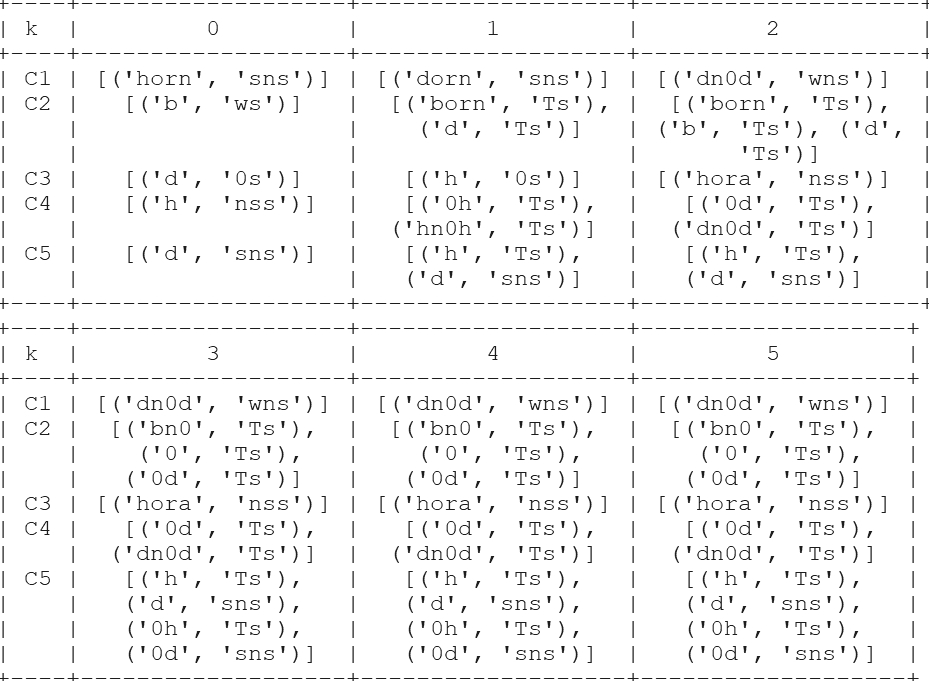}
\caption{Full map iterations with joins in formulas; insolation decreases, wind increases}
\label{fig8}       
\end{figure}

We obtain the following results for multi-valued sets as weights and (\ref{eq16}) Fig. \ref{fig8}.
One can see that confidence increases in accordance with Sec. \ref{sec2}.

\subsection{Results with the sum in formulas}
\begin{figure}
  \includegraphics[scale=1.2]{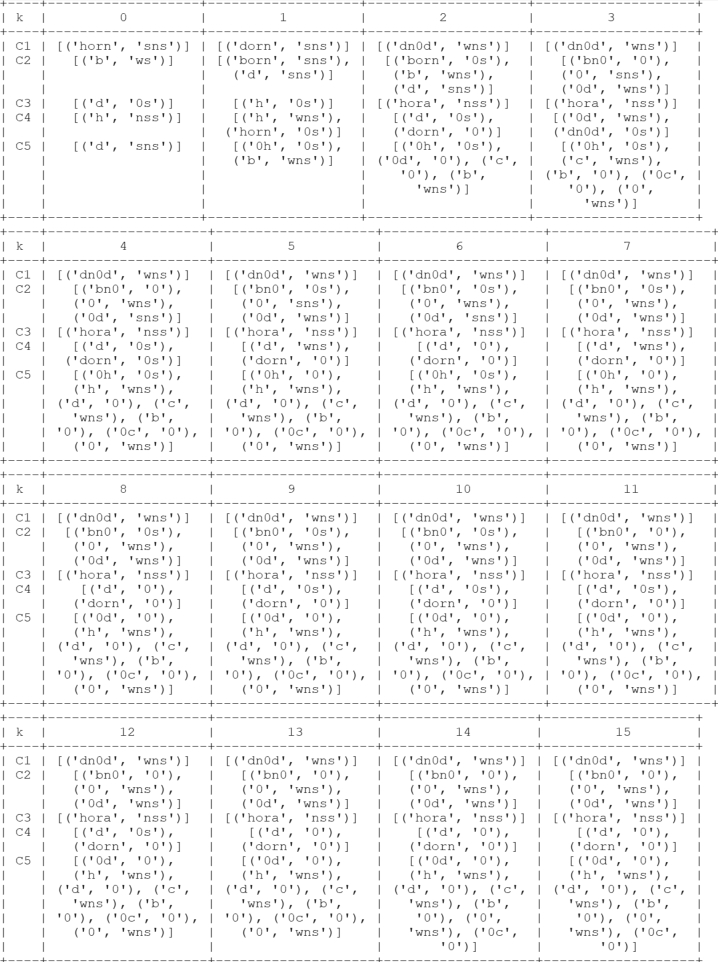}
\caption{Full map iterations with sums in formulas; insolation decreases, wind increases}
\label{fig9}       
\end{figure}

We obtain the following results for multi-valued sets as weights and (\ref{eq17}) Fig. \ref{fig9}.
Such lengthy calculations are needed to avoid looping.

\subsection{Results with mean values}
\begin{figure}
  \includegraphics[scale=0.9]{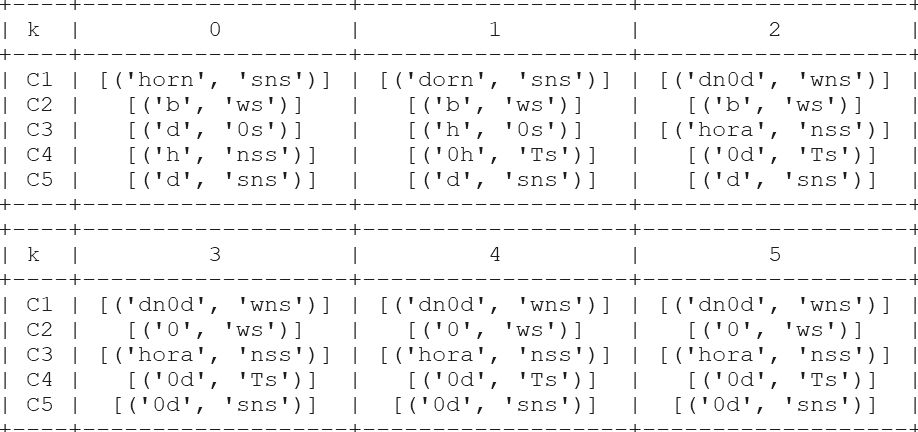}
\caption{Map iterations with the pessimistic mean of weights; insolation decreases, wind increases}
\label{fig10}       
\end{figure}

\begin{figure}
  \includegraphics[scale=0.9]{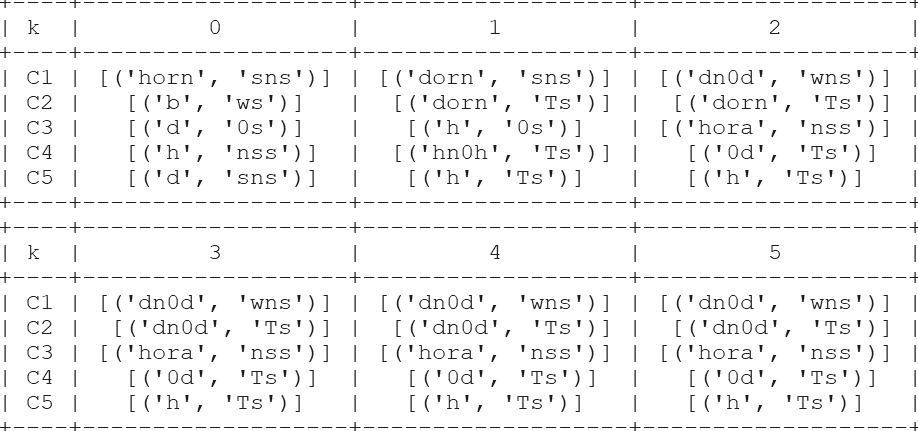}
\caption{Map iterations with the optimistic mean of weights; insolation decreases, wind increases}
\label{fig11}       
\end{figure}

We obtain the following results for the pessimistic aggregation of weights Fig. \ref{fig10} and for the optimistic aggregation of weights Fig. \ref{fig11}.

\subsection{Results with mean values and the sum in formulas}
\begin{figure}
  \includegraphics[scale=0.9]{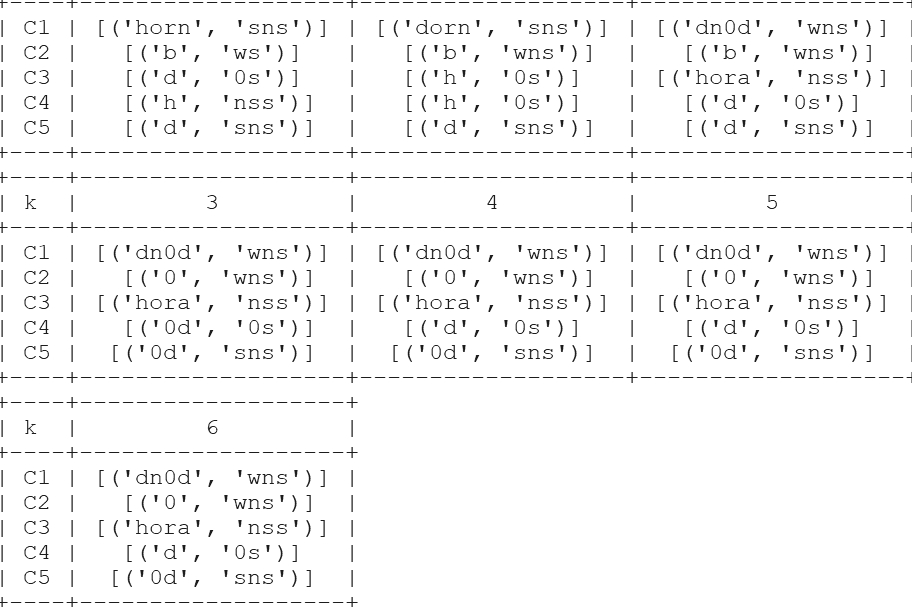}
\caption{Map iterations with the pessimistic mean of weights and the sum in formulas; insolation decreases, wind increases}
\label{fig12}       
\end{figure}

\begin{figure}
  \includegraphics[scale=0.9]{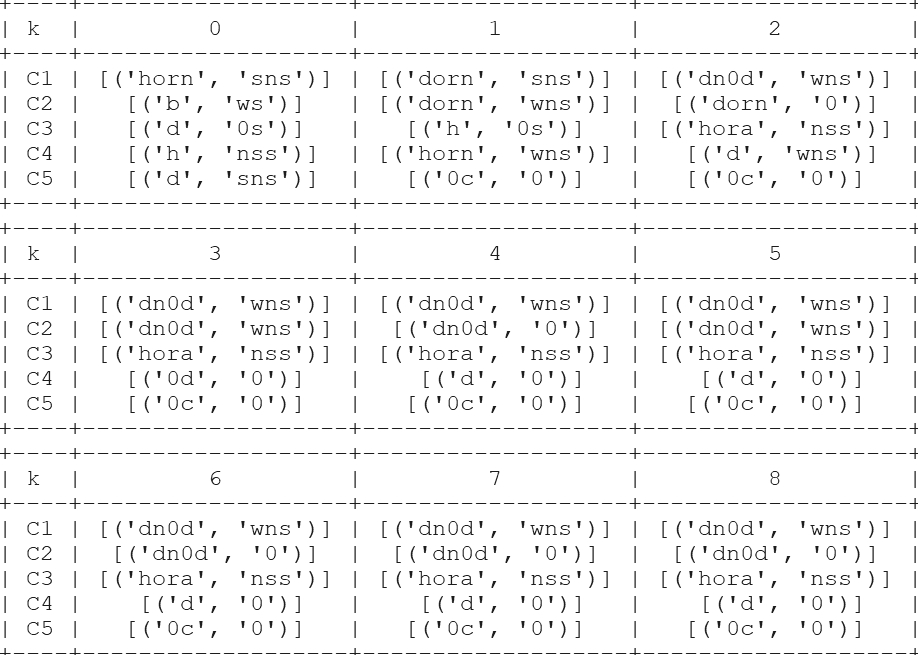}
\caption{Map iterations with the optimistic mean of weights and the sum in formulas; insolation decreases, wind increases}
\label{fig13}       
\end{figure}

We obtain the following results for the pessimistic aggregation of weights Fig. \ref{fig12} and for the optimistic aggregation of weights Fig. \ref{fig13} when (\ref{eq17}) is used.

\subsection{Discussion}
Comparing Fig. \ref{fig8} and \ref{fig9} with separate calculations in Fig. \ref{fig7}, one can see that the initial certain concept values blur into a set of several ones, i.e., the initial one-element multi-valued sets blur into real multi-valued sets. Moreover, they blur more strongly with the sum in the formulas, and they may not include the separate values in this case. Also, confidence degrees are less in the sum case, since in this case meets are subtracted from joins.

Comparing Fig. \ref{fig10} for the pessimistic mean and Fig. \ref{fig8}, you can see that the resulting concepts in  Fig. \ref{fig10} are the averages of those in Fig. \ref{fig8}. However, this is not always true for the confidence degrees, since the path to obtain final values in Fig. \ref{fig10} and Fig. \ref{fig8} are different.

For the optimistic mean, we cannot even say that the final concepts in Fig. \ref{fig11} are the averages of those in Fig. \ref{fig8} ($C_{4}$ drops out). Thus, means are not the averages of final values in general due to the difference in the paths for full and mean calculations.

Everything changes in the case of the sum: Comparing Fig. \ref{fig12} and Fig. \ref{fig13} for means and Fig. \ref{fig9}, we can obtain a mean that is not the average of the set in Fig. \ref{fig9}. It may not even be included in this set, though be of the same level (e.g., final values of $C_{5}$ Fig. \ref{fig12}).

Thus, these examples show that the formulas with join are preferable as more adequate. Other variants of a ring structure in lattices should be investigated to confirm the conclusion about the sum operation in these formulas.

\section{Conclusion}
\label{con}
In this paper, the multi-valued generalization of the concept of fuzzy number has been considered when the carrier set and the scale of confidence are lattices of a certain non-chain type. We used confidence degrees here instead of membership ones as more suitable in this case. Ordinary form of Zadeh's extension principle does not work for such a generalization, hence the principle has also been adjusted.
We also considered various options for possible definitions of the notion of mean as different similarity measures.

The illustrative example of using such multi-valued assessments in cognitive maps shows that using the join operation in cognitive map formulas is preferable to the summation operation. However, this conclusion is made only on the example of a specific sum definition in the ring of subsets of a finite set. Thus, further studies with other rings are needed. Nevertheless, the example has shown that multi-valued sets as estimators can be handled in the same way as single-valued values in the case of lattice variables, and the results are reasonable.

\appendix
\label{app}

\section{Lattices \cite{birk}}
\begin{definition}
A \textbf{lattice} is a partially-ordered set having, for any two elements, their least upper bound or join $\vee$ (sup, max) and the greatest lower bound or meet $\wedge$ (inf, min).
\end{definition}
\begin{definition}
The \textbf{least upper bound} of a subset $X\subseteq P$ of a partially-ordered set $P$ is the smallest $P$-element $a$, larger than all the elements of $X$: $min(a)\in P :\;a\geq x,\;\forall x\in P$.
\end{definition}
\begin{definition}
The \textbf{greatest lower bound} is dually defined as the largest $P$-element, smaller than all the elements of $X$.
\end{definition}
\begin{definition}
A \textbf{complete lattice} is a lattice in which any two subsets have a join and a meet.
This means that in a non-empty complete lattice there is the largest ``$\top$'' and the smallest ``0'' elements.
\end{definition}
If we take such a lattice as a scale of truth values in a multi-valued logic,
then the largest element will correspond to complete truth (true), the smallest to complete falsehood (false), and intermediate elements will correspond to partial truth in the same way as  the elements of the segment [0,1] evaluate partial truth in fuzzy logic.

In logics, with such a scale of truth values, implication can be determined by multiplying lattice elements, or internally, only from lattice operations.
\begin{definition}
Lattice elements, from which all the others are obtained
by join and meet operations are called \textbf{generators} of the lattice.
\end{definition}
\begin{definition}
A lattice is called \textbf{atomic} if every two of its generators have null meets.
\end{definition}
\begin{definition}
A \textbf{Brouwer lattice} is a lattice that has internal implications.
\end{definition}
\begin{definition}\label{imp}
In such a lattice, the \textbf{implication} $c = a\Rightarrow b$ is defined as the largest $c:\;a\wedge b = a\wedge c$.
\end{definition}
Distribution laws for join and meet are satisfied in Brouwer lattices. The converse is true only for finite lattices.

\section{Residuated Lattices \cite{resid}}\label{a2}
In non-distributive lattices, the implication cannot be defined. However, we may introduce a multiplication of the lattice elements and use it to define an external implication.
\begin{definition}
A \textbf{residuated lattice} is an algebra $(L, \vee, \wedge, \cdot, 1, \rightarrow, \leftarrow)$ satisfying the following conditions:
\begin{itemize}
\item $(L, \vee, \wedge)$ is a lattice;\\
\item $(L, \cdot, 1)$ is a monoid; \\
\item $(\rightarrow, \leftarrow)$ is a pair of residuals of the operation $\cdot$, that means
\begin{equation*}
\forall x, y \in L: x\cdot y \leqslant z \Leftrightarrow y\leqslant x\rightarrow z \Leftrightarrow x\leqslant z\leftarrow y
\end{equation*}
\end{itemize}
In this case, the operation $\cdot$ is order preserving in each argument and for all $a, b \in L$ both the sets $\{y\in L| a\cdot y \leqslant b\}$ and $\{x\in L| x\cdot a \leqslant b\}$ each contains a greatest element ($a\rightarrow b$ and $b\leftarrow a$ respectively).
\end{definition}
The monoid multiplication $\cdot$ is distributive over $\vee$:
\begin{equation*}
x\cdot (y\vee z) = (x\cdot y)\vee(x\cdot z).
\end{equation*}
Also, $x\cdot 0 = 0\cdot x = 0$. A special case of residuated lattices is a Heyting algebra, when the monoid multiplication coincides with $\wedge$.

In non-commutative monoids, residuals $\rightarrow$ and $\leftarrow$ can be understood as having a temporal quality: $x\cdot y \leqslant z$ means ``$x \;then\; y \;entails\; z$,'' $y \leqslant x\rightarrow z$ means ``$y \;estimates\; the\; transition \;had\; x\; then\; z$,'' and $x \leqslant z\leftarrow y$ means ``$x\; estimates\; the\; opportunity$ \emph{if-ever} $y\; then\; z$.'' You may think about $x, y$, and $z$ as $bet,\; win$, and $rich$ correspondingly (Wikipedia).

\begin{definition}
A residuated lattice A is said to be \textbf{integrally closed} if it satisfies the equations
$x\cdot y \leqslant x \Longrightarrow y \leqslant 1$ and $y\cdot x \leqslant x \Longrightarrow y \leqslant 1$, or equivalently, the equations $x\rightarrow x = 1$ and $x\leftarrow x = 1$ \\ \cite{integr}.
\end{definition}
Any upper or lower bounded integrally closed residuated lattice $L$ is \textbf{integral}, i.e., $a\leqslant 1,\;\forall a\in L$ \cite{integr}.

\section*{References}

  \bibliographystyle{elsarticle-num}
  \bibliography{Maximov_Generalizations}




\end{document}